\documentclass{article} 
\usepackage{iclr2026_conference,times}

\usepackage{amsmath,amsfonts,bm}

\def\eqref#1{equation~\ref{#1}}

\def\1{\bm{1}}

\DeclareMathAlphabet{\mathsfit}{\encodingdefault}{\sfdefault}{m}{sl}
\SetMathAlphabet{\mathsfit}{bold}{\encodingdefault}{\sfdefault}{bx}{n}

\usepackage[hidelinks]{hyperref}
\usepackage{url}
\usepackage{graphicx}
\usepackage{booktabs}
\usepackage{multirow}
\usepackage{subcaption}
\usepackage{siunitx}
\usepackage{enumitem}
\usepackage{microtype}
\usepackage{caption}
\title{ProcessThinker: Enhancing Multi-modal Large Language Models Reasoning via Rollout-based Process Reward}

\author{
Jingpei Wu$^{1,5}${\NoHyper\thanks{Equal contribution.}\endNoHyper} \quad
Xiao Han$^{1}${\NoHyper\footnotemark[1]\endNoHyper} \quad
Weixiang Shen$^{1}$ \quad
Boer Zhang$^{2}$ \quad
Zifeng Ding$^{3,4}$ \quad
Volker Tresp$^{1,5}$ \\
$^{1}$LMU Munich \quad 
$^{2}$Harvard University \quad
$^{3}$University of Cambridge \quad
$^{4}$Mina AI \\
$^{5}$Konrad Zuse School of Excellence in Reliable AI (relAI) \quad
}

\iclrfinalcopy 
\begin{document}

\maketitle
\lhead{Published at ICLR 2026 Workshop on Logical Reasoning of Large Language Models}
\begin{abstract}
Visual question answering increasingly requires multi-step reasoning. Recent post-training with reinforcement learning under verifiable rewards (RLVR) and Group Relative Policy Optimization (GRPO) can improve multimodal reasoning, but most approaches rely on sparse outcome-only rewards. As a result, they struggle to tell whether an incorrect answer comes from a small mistake late in the reasoning or from an unhelpful trajectory from the start. A common solution is to train a process reward model (PRM) for step-level supervision, but this typically requires large-scale high-quality chain-of-thought annotations and additional training cost. We propose \textbf{ProcessThinker}, a practical post-training pipeline that provides step-level \emph{process rewards} without training an explicit PRM. ProcessThinker first rewrites reasoning traces into a step-tagged format for cold-start supervised fine-tuning, then applies GRPO with a standard format reward and our rollout-based process reward. Concretely, for each intermediate step, we sample multiple continuations from that step and use the empirical success rate (final-answer verification) as the step reward. This gives dense credit assignment and encourages reasoning steps that more reliably support a correct conclusion, helping reduce inconsistent or self-contradictory progress across steps -- a key issue in logical reasoning. Across four challenging video benchmarks (Video-MMMU, MMVU, VideoMathQA, and LongVideoBench), ProcessThinker consistently improves over the baseline model \textsc{Qwen3-VL-8B-Instruct}.
\end{abstract}

\section{Introduction}
\label{sec:intro}
Multimodal large language models (MLLMs) have made rapid progress in open-ended visual understanding and question answering. With chain-of-thought (CoT) prompting, MLLMs can produce multi-step reasoning traces and often improve task performance. This long-horizon reasoning ability is increasingly important for complex problems where the answer depends on a sequence of intermediate inferences.
Recently, post-training with reinforcement learning under verifiable rewards (RLVR), often paired with group-based objectives such as Group Relative Policy Optimization (GRPO), has further strengthened multi-step reasoning in both text-only and multimodal settings~\citep{shao2024deepseekmath,deepseek2025r1,su2025rlvrdomain,sim2025groundedrlvr,feng2025video,zhang2025r1vl,park2025deepvideor1,wang2025timer1,zhang2025tinyllava,yang2025r1onevision,huang2025visionr1}. 
However, the supervision signal in GRPO-style RLVR is typically \emph{sparse}: the verifier only checks the final answer. For long reasoning traces, many samples within a group can receive identical outcome rewards, which weakens learning signals and motivates reward shaping and sampling strategies~\citep{yao2025sharegrpo,xu2025xrpo,zhang2025grpo_lead,niu2026rlrr,yari2026amirgrpo,tao2025hero,lyu2025oreal}. Even with these improvements, outcome-only supervision still provides little information about \emph{which intermediate steps} were helpful when the final answer is wrong.

A natural way to densify supervision is to score intermediate steps. Process reward models (PRMs) provide step-level feedback and have been used for reranking, search, and test-time scaling by evaluating the quality of the reasoning process~\citep{lightman2023verify,setlur2024rewardingprogress,khalifa2025thinkprm,zhao2025genprm,wang2025visualprm,du2025mmprm}. 
Recent work also explores using PRMs to supply process rewards during RL training~\citep{luo2025unlocking}. 
However, PRM-based supervision usually requires high-quality step annotations (or a complex pipeline to synthesize them), and automated approaches often rely on Monte-Carlo rollouts or MCTS-style search, which can be noisy and sensitive to how ``steps'' are defined~\citep{zhang2024restmcts,zhang2025rest_rl,zhang2025lessonsprm,tan2025aurora,ding2025scan}. Moreover, training and maintaining a separate PRM adds engineering overhead and can introduce a mismatch between the PRM and the final policy.

\begin{figure}[t]
    \centering
    \includegraphics[width=0.9\textwidth]{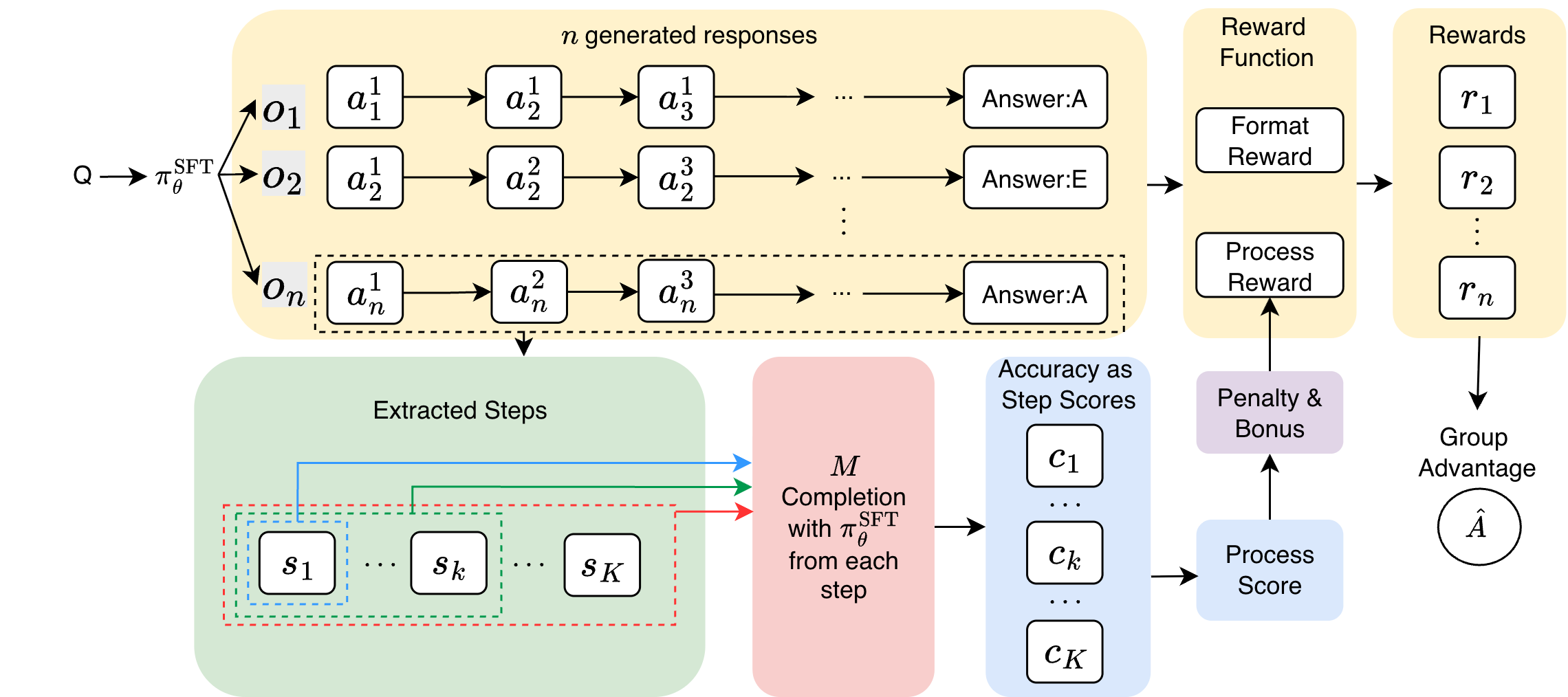}
    \caption{Rollout-based process reward inside one GRPO update. For a question $Q$, we sample a group of $G$ candidate responses from the current policy. For each response, we extract step segments $\{s_t\}$ (incl. multiple $a$) and score each step $k$ by the success rate of $M$ continuation rollouts from previous steps $(s_1, \dots, s_k)$ under policy model, producing step scores $c_t$ and an averaged process score. The final reward for each response combines format reward, process reward, and step-count shaping (bonus + penalty gate), which is then used to compute group-relative advantages for GRPO.}
    \label{fig:pipeline}
\end{figure}

This raises a question central to logical multi-step reasoning: \emph{can we obtain step-level training signals that encourage more consistent reasoning traces \textbf{without} training a separate PRM?}
StepGRPO~\citep{zhang2025r1vl} is an important step in this direction, using rule-based step-wise rewards (e.g., rewarding the presence of key steps and enforcing a well-structured reasoning format). However, it still does not directly measure whether a specific intermediate step actually makes the problem easier to solve.
VinePPO~\citep{kazemnejad2025vineppo} shares a similar intuition, using Monte Carlo rollouts to estimate step-level values for credit assignment in PPO; however, it targets text-only LLMs and does not provide rollout-based process rewards within a GRPO framework.
We propose \textbf{ProcessThinker}, which assigns a rollout-based \emph{process reward} to each reasoning step via \emph{continuation solvability} (Figure~\ref{fig:pipeline}). 
The key idea is simple: an intermediate step is useful if, conditioned on the partial trace, the model is more likely to reach the correct final answer. We estimate this by sampling multiple continuations from the current policy starting from each step prefix and computing the empirical success rate under the same final-answer verifier used in RLVR. This provides a direct, model-free estimate of step utility and encourages reasoning traces whose steps more reliably support a correct conclusion, reducing inconsistent progress across steps---a core challenge in logical reasoning~\citep{lightman2023verify}.
We instantiate ProcessThinker on \textsc{Qwen3-VL-8B-Instruct}~\citep{bai2025qwen3vl} and train in two stages: 
(i) a SFT warm-up on a step-tagged dataset obtained by rewriting \textsc{Video-R1-CoT} traces into explicit step decompositions using a stronger teacher model, and 
(ii) GRPO post-training with a weighted combination of sparse outcome reward and our rollout-based process reward, along with lightweight formatting incentives.
We evaluate in the video domain, but the proposed reward construction is model- and modality-agnostic.

In summary, we make the following contributions:
(1) We propose a simple GRPO-based post-training framework that incorporates step-level rewards \emph{without} training an explicit process reward model (PRM). 
(2) We introduce a rollout-based process reward that scores each reasoning step by the empirical success rate of multiple continuations conditioned on the step prefix.
(3) We demonstrate consistent improvements over \textsc{Qwen3-VL-8B-Instruct} on four video reasoning benchmarks, and ablations show that increasing the weight of the process reward yields larger gains.

\section{Method}
\label{sec:method}

Given a multimodal context $x$ (video frames or an image plus text prompt), the model generates an output $y$.
We enforce a step-tagged reasoning format
\begin{equation}
y=\langle \texttt{think}\rangle \dots \langle \texttt{step}\rangle s_k \langle/\texttt{step}\rangle \dots \langle/\texttt{think}\rangle \langle \texttt{answer}\rangle ans \langle/\texttt{answer}\rangle ,
\end{equation}
which makes intermediate steps explicit and enables step-wise scoring (Figure~\ref{fig:pipeline}).

\subsection{SFT data construction (format + filtering)}
Starting from \textsc{Video-R1-CoT-165k}~\citep{feng2025video}, we rewrite each sample into the step-tagged format using a stronger teacher model \textsc{Qwen3-VL-30B-A3B-Instruct}~\citep{bai2025qwen3vl}.
The teacher is instructed to preserve the original solution while segmenting the reasoning into non-trivial, non-redundant steps.
To reduce rewriting noise (missing/duplicated steps, semantic drift, step-answer mismatch), we apply a second-pass filter with the teacher that scores: (i) answer fidelity vs.\ the original solution, (ii) consistency between steps and the final answer, and (iii) step quality. We keep the top 19k samples for SFT and sample 1{,}250 prompts for RL.
We fine-tune \textsc{Qwen3-VL-8B-Instruct} on the 19k set to obtain ProcessThinker-SFT, which reliably emits parsable step-tagged traces.

\subsection{GRPO with rollout-based process rewards}
\label{subsec:grpo}
\noindent \textbf{GPRO} For each prompt $x$, we sample a group of $G$ responses $\{y^{(g)}\}_{g=1}^G$ from the current policy $\pi_\theta(\cdot|x)$, compute a scalar reward $r^{(g)}$ for each response, and normalize rewards within the group to obtain relative advantages (mean/variance normalization). The policy is then updated using the standard GRPO recipe with KL regularization to a reference policy.
ProcessThinker differs from prior RLVR work mainly in the reward design below.

\noindent \textbf{Rollout-based process reward (continuation solvability).}
For a sampled response $y$ with steps $s_{1:K}$ and ground-truth answer $ans^\star$, we score each prefix $p_i=(s_1,\ldots,s_i)$ by how often the model can successfully \emph{finish} the problem when conditioned on that prefix.
We sample $M$ continuations $\hat y_i^{(m)}\sim\pi_\theta(\cdot|x,p_i)$ and define the step score as the empirical success rate:
\begin{equation}
c_i=\frac{1}{M}\sum_{m=1}^{M}\mathbf{I}\!\left[\mathrm{Ans}(\hat y_i^{(m)})=ans^\star\right].
\end{equation}
The trajectory-level process reward averages prefix solvability:
\begin{equation}
R_{\text{proc}}(y)=\frac{1}{\min(K,K_{\max})}\sum_{i=1}^{\min(K,K_{\max})}c_i,
\end{equation}
using $M{=}4$ and $K_{\max}{=}6$ unless noted. This gives dense credit assignment: early steps can receive partial credit even if the final answer in $y$ is wrong.

\noindent \textbf{Format reward, bounded step bonus, and penalty gate.}
We use a strict format reward $r_{\text{fmt}}$ that is awarded only if tags are properly nested and $K\in[K_{\min},K_{\max}]$ (and optionally a length bonus if within $[L_{\min},L_{\max}]$), similar in spirit to step-structured RL recipes.
To encourage using more than the minimum number of steps without step inflation, we add a bounded step bonus
\begin{equation}
B(K)=\alpha \sqrt{\mathrm{clip}\Big(\frac{K-K_{\min}}{K_{\max}-K_{\min}},\,0,\,1\Big)}.
\end{equation}
To reduce reward hacking (too few steps or shallow steps that only satisfy format), we gate rewards with a simple penalty:
\begin{equation}
\bar R_{\text{acc}}=
\begin{cases}
1, & R_{\text{acc}}=1,\\
-\,B(K), & \text{otherwise},
\end{cases}
\qquad
\bar R_{\text{proc}}=
\begin{cases}
R_{\text{proc}}, & R_{\text{proc}}\ge\tau,\\
-\,B(K), & \text{otherwise},
\end{cases}
\end{equation}
where $R_{\text{acc}}\in\{0,1\}$ is final-answer accuracy and $\tau{=}0.5$.

\noindent \textbf{Final reward.}
For format-valid responses, the reward is
\begin{equation}
r^{(g)}=(r_{\text{fmt}}+\beta)\;+\;\lambda_{\text{acc}}\bar R_{\text{acc}}
\;+\;\lambda_{\text{proc}}\bar R_{\text{proc}}\;+\;B(K),
\qquad \lambda_{\text{acc}}+\lambda_{\text{proc}}=1.
\label{eq:final_reward}
\end{equation}
If formatting is invalid, we set $r^{(g)}{=}0$ and skip continuation rollouts for efficiency.

\section{Experiments}

\begin{table*}[t]
\vspace{-4pt}
\centering
\small
\setlength{\tabcolsep}{2pt}
\renewcommand{\arraystretch}{1.0}
\sisetup{table-format=2.2, table-number-alignment=center}
\begin{tabular}{l S S S S S}
\toprule
\multirow{2}{*}{\textbf{Model}} & \multicolumn{5}{c}{\textbf{Accuracy (\%) $\uparrow$}} \\
\cmidrule(lr){2-6}
 & {Video-MMMU} & {MMVU (mc)} & {VideoMathQA} & {LongVideoBench} & {Avg.} \\
\midrule
\textsc{Qwen3-VL-8B-Instruct} & 62.89 & 65.60 & 25.20 & 71.50 & 56.30 \\
\textsc{Video-R1-7B}\citep{feng2025video} & 53.89 & 65.92 & 26.67 & 58.30 & 51.20 \\
\textsc{ProcessThinker-SFT} & 58.78 & 64.48 & 23.57 & 68.50 & 53.83 \\
\midrule
ProcessThinker (outcome-only) & 60.78 & 67.36 & 27.86 & 74.20 & 57.55 \\
ProcessThinker (outcome + process) & 61.67 & 67.52 & 27.86 & 74.60 & 57.91 \\
\textbf{ProcessThinker (process-only)} & \textbf{63.33} &  \textbf{68.48} & \textbf{31.67} & \textbf{75.40} & \textbf{59.72} \\
\bottomrule
\end{tabular}
\caption{Main results on four video reasoning benchmarks.
All ProcessThinker variants share the same SFT warm-up and are trained with GRPO using the overall reward in ~\eqref{eq:final_reward}.
``process-only'' uses our rollout-based step-wise process reward (Section~\ref{subsec:grpo}), while ``outcome-only'' uses only final-answer correctness.}
\label{tab:main_results}
\vspace{-4pt}
\end{table*}

We evaluate on four video reasoning benchmarks: \textsc{Video-MMMU}~\citep{hu2025videommmu}, \textsc{MMVU}~\citep{zhao2025mmvu}, \textsc{VideoMathQA}~\citep{rasheed2025videomathqa}, and \textsc{LongVideoBench}~\citep{wu2024longvideobench}, and report accuracy following each benchmark's official protocol.

\textbf{Training setup.}
All ProcessThinker variants share the same SFT warm-up and differ only in the reward mixture $(\lambda_{\text{acc}},\lambda_{\text{proc}})$ in ~\eqref{eq:final_reward}.
Unless otherwise stated, we sample $G{=}4$ responses per prompt for GRPO and compute process rewards with $M{=}4$ continuation rollouts per step, capped at $K_{\max}{=}6$.

\textbf{Main results.}
As shown in Table~\ref{tab:main_results}, ProcessThinker (process-only) improves over the \textsc{Qwen3-VL-8B-Instruct} baseline on all four benchmarks, raising the average score from 56.30 to 59.72 (+3.42).
\textsc{Video-R1-7B}~\citep{feng2025video} is included for reference, though it uses the older \textsc{Qwen2.5-VL} backbone and is not directly comparable.
The largest gain is on \textsc{VideoMathQA} (+6.47), which requires multi-step reasoning while integrating sparse cues over time; we also observe a clear improvement on \textsc{LongVideoBench} (+3.90), suggesting better reasoning under long-context inputs. 
Overall, the consistent gains across benchmarks indicate that step-level credit assignment from rollout-based process rewards transfers beyond a single dataset type.

\textbf{SFT warm-up: necessary but insufficient.}
\textsc{ProcessThinker-SFT} underperforms the instruction-tuned baseline despite improved format compliance, suggesting that step-tag supervision mainly teaches \emph{how to write} structured traces, but does not directly optimize verified task success.
GRPO post-training is therefore critical to turn well-formed reasoning traces into higher final-answer accuracy.

\textbf{Process reward vs.\ outcome reward.}
Among reward mixtures, \emph{process-only} performs best, while \emph{outcome-only} and a balanced mixture lag behind.
This supports our hypothesis that outcome-only supervision is too sparse for long-horizon reasoning, whereas rollout-based process rewards provide denser signals that better capture the usefulness of intermediate steps.

\section{Conclusion}
ProcessThinker shows that step-wise credit assignment can be obtained from sparse verifiable outcomes without training a separate PRM: we score each step by its \emph{continuation solvability} via rollout success rates. This rollout-based process reward encourages intermediate steps that more reliably support a correct conclusion, addressing a key issue in logical multi-step reasoning: unproductive or inconsistent progress across steps.
Combined with strict formatting constraints and GRPO-style RLVR, ProcessThinker consistently improves \textsc{Qwen3-VL-8B-Instruct} on four video reasoning benchmarks, with the strongest gains when emphasizing process rewards over outcome-only supervision. 
The main limitation is efficiency: although we avoid PRM annotation/training, continuation-solvability requires multiple rollouts per step, and the added inference cost can offset the benefit of using fewer data. The resulting reward can also be noisy due to rollout stochasticity and sensitivity to step segmentation.

\subsubsection*{Acknowledgments}
This paper is supported by the DAAD programme Konrad Zuse Schools of Excellence in Artificial Intelligence, sponsored by the Federal Ministry of Research, Technology and Space.

\bibliography{iclr2026_conference}
\bibliographystyle{iclr2026_conference}


\end{document}